\documentclass[letterpaper, 10 pt, conference]{ieeeconf}

\IEEEoverridecommandlockouts                              %

\let\labelindent\relax

\usepackage{tikz}
\usepackage{graphicx} %
\usepackage{subcaption}
\usepackage{cuted}
\usepackage{booktabs}
\usepackage{multirow}
\usepackage{enumerate}
\usepackage{enumitem}
\usepackage{microtype}
\usepackage{algorithm}
\usepackage{algpseudocode}
\usepackage{cite}
\usepackage{siunitx}
\usepackage[breaklinks,colorlinks]{hyperref}
\usepackage[hang,flushmargin,symbol]{footmisc}
\usepackage{makecell}
\usepackage{balance}

\usepackage{amsmath}
\usepackage{amssymb}
\usepackage{tensor}

\DeclareMathAlphabet{\mathcal}{OMS}{cmsy}{m}{n}

\def\R{\ensuremath{\mathbb{R}}}

\renewcommand{\vec}[1]{\ensuremath{#1}}

\newcommand{\norm}[1]{\left\lVert#1\right\rVert}

\newcommand{\cardinality}[1]{\left|#1\right|}

\DeclareMathOperator*{\argmax}{arg\,max}
\DeclareMathOperator*{\argmin}{arg\,min}

\usepackage{xspace}

\newcommand{\ie}{\mbox{i.\,e.}\xspace}

\newcommand{\etal}{\emph{et al.}\xspace}
   
\renewcommand{\[}{\begin{equation}}
\renewcommand{\]}{\end{equation}}

\renewcommand{\baselinestretch}{0.99}

\usepackage[capitalize]{cleveref}

\crefname{figure}{Fig.}{Figs.}
\Crefname{figure}{Figure}{Figures}
\crefname{section}{Sec.}{Secs.}
\Crefname{section}{Section}{Sections}
\Crefname{table}{Table}{Tables}
\crefname{table}{Tab.}{Tabs.}
\crefname{algorithm}{Algo.}{Algos.}
\Crefname{algorithm}{Algorithm}{Algorithms}
\crefname{appendix}{Sec.}{Secs.}
\Crefname{appendix}{Section}{Sections}

\title{The Neural Compass: Probabilistic Relative Feature Fields for Robotic Search}
\author{Gabriele Somaschini, Adrian Röfer, and Abhinav Valada
\thanks{\noindent The authors are with the Department of Computer Science, University of Freiburg.}
\thanks{\noindent This work was funded by the BrainLinks-BrainTools center of the University of Freiburg.}
}

\def\GT{\text{GT}}

\makeatletter
\g@addto@macro{\endtabular}{\rowfont{}}%
\makeatother
\newcommand{\rowfonttype}{}%
\newcommand{\rowfont}[1]{%
\gdef\rowfonttype{#1}#1\ignorespaces
}

\newcommand{\ours}{ProReFF}

\newif\ifmix
\mixtrue %

\begin{document}
\bstctlcite{IEEEexample:BSTcontrol} %
\maketitle
\pagestyle{empty}

\begin{abstract}
    Object co-occurrences provide a key cue for finding objects successfully and efficiently in unfamiliar environments. Typically, one looks for cups in kitchens and views fridges as evidence of being in a kitchen. Such priors have also been exploited in artificial agents, but they are typically learned from explicitly labeled data or queried from language models. It is still unclear whether these relations can be learned implicitly from unlabeled observations alone. In this work, we address this problem and propose \ours{}, a feature field model trained to predict \emph{relative} distributions of features obtained from pre-trained vision language models. In addition, we introduce a learning-based strategy that enables training from unlabeled and potentially contradictory data by aligning inconsistent observations into a coherent relative distribution.
    For the downstream object search task, we propose an agent that leverages predicted feature distributions as a semantic prior to guide exploration toward regions with a high likelihood of containing the object. We present extensive evaluations demonstrating that \ours{} captures meaningful relative feature distributions in natural scenes and provides insight into the impact of our proposed alignment step. We further evaluate the performance of our search agent in 100 challenges in the Matterport3D simulator, comparing with feature-based baselines and human participants. The proposed agent is $20\%$ more efficient than the strongest baseline and achieves up to $80\%$ of human performance. 
\end{abstract}

\section{Introduction}
\label{section:intro}

One of the fundamental challenges for household robots is localizing
objects in previously unseen environments.
Humans possess strong priors about the structure of domestic spaces. Even in an unfamiliar home, one naturally heads to the kitchen to look for a cup rather than the bathroom, and is more likely to look for the missing TV remote on the sofa than on the dining table.
This concept has been coined \emph{object co-occurrences}. %
As these priors enable efficient exploration, robotic agents should also be endowed with them.

Prior work has studied obtaining such priors from internet data, or annotated datasets~\cite{kollar2009utilizing,pronobis,zeng2020semantic,prasanna2024perception}. Given these priors, agents can explore spaces efficiently by scoring which regions to prioritize~\cite{schmalstieg2023learning, schmalstieg2022learning}. In recent years, Large Language Models (LLMs) have become a new, powerful source of prior knowledge, enabling the search for arbitrary objects by leveraging scene graph structures to make the search manageable~\cite{honerkamp2024momallm,yin2024sg,loo2025open, mohammadi2025more}.
However, the navigation remains tied to object instances and being able to make proposals for them. Moreover, this also requires online scene graphs construction.

Recent advances in self-supervised visual representation learning have also enabled a different set of methods that use the similarity between a current observation and a goal feature to decide on an exploration direction~\cite{gadre2023cows,zhou2023esc,yokoyama2024vlfm}.
These methods build on models such as DINOv2~\cite{oquab2023dinov2} and CLIP~\cite{radford2021learning}, which produce rich, general-purpose patch-level features from unlabeled images, yielding a continuous, open-vocabulary embedding space in which any object can be represented without prior labeling.
Prior work has demonstrated that such features can be meaningfully encoded in neural feature fields, enabling open-vocabulary grasp selection and language queries for regions in 3D scenes~\cite{kobayashi2022decomposing,kerr2023lerf,shen2023distilled}.

\begin{figure}
    \centering
    \vspace{2mm}
    \includegraphics[width=0.9\linewidth]{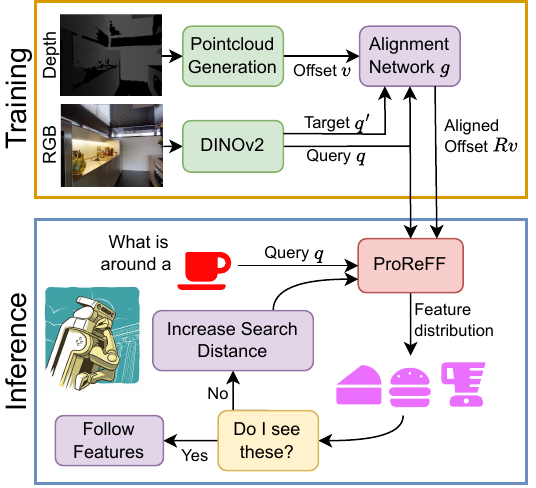}
    \caption{Overview of \ours{}. We train a relative feature field, which predicts a mean embedding and a variance, given a query embedding $\vec{q}$ and a relative offset $\vec{v}$. This network is trained in an unsupervised manner from feature point cloud observations. To enable this training, we introduce a learned data alignment model. We demonstrate the utility of \ours{} for object search by using it to infer distributions of features around a target object. These distributions are compared with the agent's current observations, and a choice is made between following the current observations or inferring further features.}
    \label{fig:model}
\end{figure}

In this work, we address the problem of estimating object co-occurrences in a self-supervised manner. Instead of learning explicit co-occurrences or querying for specific object relations, we employ a self-supervised method to learn a feature field directly from the co-occurrences of low-level vision-language features. 
We coin this representation \emph{Probabilistic Relative Feature Fields} (\ours{}).
Rather than reconstructing a specific scene, \ours{} encodes the statistical co-occurrence structure across environments. Given a semantic feature query and spatial offsets relative to the query, \ours{} predicts the distribution of features likely to appear at those relative locations.
We present an exploration strategy that uses \ours{} to consider different probable spatial context of a target object and guides exploration toward semantically promising regions. We evaluate it against different baselines and human participants on the Matterport3D~\cite{chang2017matterport3d} dataset.
To the best of our knowledge, this is the first use of a feature field to encode cross-environment spatial priors for object navigation. Our specific contributions are:
\begin{itemize}
    \item \ours{}: a probabilistic feature field encoding spatial
    co-occurrence structure across environments and trained in a fully self-supervised manner without semantic labels.
    \item An object search strategy guided by \ours{} utilizing different scales of semantic spatial context.
    \item Evaluation on Matterport3D contrasting with other zero-shot object search agents and human participants on the object search task.
    \item We release the code and trained models upon acceptance.
\end{itemize}

\section{Related Work}

\subsection{Object Search \& Co-Occurrences}

In \emph{ObjectNav}, an agent is tasked with finding an object of a target category in unseen environments~\cite{batra2020objectnav}. Fundamental to this challenge is the exploration of \emph{frontiers}, the boundaries between explored and unexplored space~\cite{yamauchi1997frontier}. To successfully and efficiently find an object, one needs a scoring function to decide which frontier should be explored.
Typically, these scoring functions build in the idea of \emph{object co-occurrences}, \ie, cups are more likely to be found near a fridge than a garage.
Classically, such co-occurrences have been obtained from annotated datasets, or mined from internet data and integrated into probabilistic planners~\cite{aydemir2011search,pronobis,kollar2009utilizing,zeng2020semantic}.
More recently, approaches such as SemExp~\cite{chaplot2020object} build explicit semantic maps which record the presence of different object categories as one-hot vectors. Given such maps, a learned high-level exploration agent selects areas to explore.
All of these approaches represent object categories distinctively, making transfer to novel categories challenging without additional annotated data.

The recent rise of Vision Language Models (VLMs) has enabled a new class of approaches that are capable of open-vocabulary zero-shot navigation.
Approaches such as CoW~\cite{gadre2023cows} combine CLIP~\cite{radford2021learning}-based object localization with frontier exploration, matching the performance of methods trained for hundreds of millions of steps.
VLFM~\cite{yokoyama2024vlfm} and OneMap~\cite{busch2025one} generate language-grounded value maps from RGB observations to score frontiers by comparing the embedding vector of the current scene observation to that of a target text prompt.
A source of object co-occurrences that has recently gained popularity is Large Language Models (LLMs). 
ESC~\cite{zhou2023esc} improves on CoW by constraining the exploration using candidate prompts that are scored by an LLM.
In order to expose larger spaces for reasoning and not rely on the quality of a dense map, scene graphs can be built as a robotic agent progresses through an environment~\cite{hughes2022hydra,kurenkov2023modeling}.
Coupled with LLMs, such scene graph representations can be used to perform effective navigation in an open vocabulary setting by using the LLM to score graph nodes for exploration~\cite{honerkamp2024momallm,yin2024sg,loo2025open}.

Our approach similarly builds on the advantages of VLMs but seeks to reduce reliance on explicit object labels. While LLM-based approaches are open-vocabulary, they require object proposals and names to function. While this is not the case for approaches such as VLFM, they can only score the immediate direction for exploration.
By training a relative feature field, we aim to enable an approach that can predict beyond the current frontier without requiring object candidates or training labels. 

\subsection{Feature Fields \& Semantic Visual Features}

Neural fields are trained functions that predict values given 3D coordinates and additional optional queries.
Neural fields have been a popular approach for representing 3D objects or environments as Neural Radiance Fields (NeRF)~\cite{irshad2024neural}.
NeRFs are fitted using differentiable volume rendering and predict densities and radiance values at 3D coordinates. 
Beyond visual scene reproduction, several works have derived utility from Neural Fields by storing high-dimensional semantic features~\cite{kobayashi2022decomposing} within the fields rather than RGB radiance.
VLMs such as CLIP~\cite{radford2021learning} provide features for image patches, which match the vector generated for an equivalent linguistic description. Using differentiable rendering, such features have been embedded in Neural Fields, enabling scene querying~\cite{kerr2023lerf}, language-guided manipulation~\cite{shen2023distilled,weng2022neural,chen2024funcgrasp}, and scene understanding~\cite{gosala2024letsmap}.

In this work, we fit a probabilistic Neural Field in an unsupervised manner on feature point clouds generated from RGB-D observations. Unlike most Neural Fields, we do not seek to store a specific scene, but obtain a field which encodes a general notion of how features relate spatially. Different from direct scene completion, as done by~\cite{jevtic2025feed}, the generality of this field should enable inference that is also valid without the context of the current view. 
We build our approach on features obtained from DINOv2~\cite{oquab2023dinov2,darcet2023vision}. Amir~\etal~\cite{amir2021deep} empirically showed that these features encode well-localized semantic information at fine spatial granularity, making them a reliable source for our task.
While DINOv2 itself does not provide a language encoder, Talk2DINO~\cite{barsellotti2025talking} provides a learned mapping, which maps CLIP language embeddings to the DINOv2 feature space. This enables fine-grained language-based queries on DINOv2, thereby overcoming the region-based language extensions used previously~\cite{liu2024grounding}.

\section{Approach}
\label{sec:approach}

We introduce our approach in three parts. First, we present the core probabilistic relative feature field, \ours{}, followed by a learned data augmentation that enables its successful training. Lastly, we introduce our object search agent, which uses the trained \ours{}.

\subsection{Relative Feature Field Model}
\label{sec:ap-feature-field}

The aim of our method is to predict the spatially relative occurrence of deep feature embeddings. Given, for example, a feature stemming from a stove, we aim for our model to capture the possibility of a pot being found near the stove. Additionally, a bit farther away, the model should be able to indicate the presence of a fridge or a sink.  
To achieve this goal, we assume to be given a query feature $\vec{q} \in \R^E$ (where $E=768$) and a displacement vector $\vec{v} \in \R^3$ for which our model predicts a mean feature $\vec{\mu} \in \R^E$ and a scalar variance $\sigma{^2} \in \R$ representing the predicted spread of features around $\vec{\mu}$. Our full model is thus a map
\[
    f : \R^E \times \R^3 \rightarrow \R^E \times \R.
\]
We implement $f$ as an MLP with 8 layers, 256 hidden units (using ReLU activations), and a skip connection 
that concatenates the inputs to the fifth layer's activation, following the 
architecture of NeRF~\cite{mildenhall2021nerf}. Unlike NeRF, we do not apply 
positional encoding to the inputs, as this encoding is applied to obtain sharp, high-contrast fields.
This stands in opposition to our aim to fit a field that encodes general trends.
In order to train this model, we require training triplets $(q, \vec{v}, q')$ 
consisting of a query embedding, a displacement vector, and a target embedding, 
whose construction is described in \cref{sec:training}.
As our model is meant to predict distributions, we train $f_\theta$ by minimizing a cosine-based negative log-likelihood loss over all 
training pairs in $\mathcal{D}$:
\begin{equation}
    \label{eq:loss}
    \mathcal{L} = \frac{1}{|\mathcal{D}|} \sum_{(q, v, q') \in \mathcal{D}} 
    \frac{\left(1 - \cos(q',\, \mu)\right)^2}{2\sigma^2} + \frac{1}{2} \log \sigma^2,
\end{equation}
where $(\mu, \sigma^2) = f_\theta(q, v)$ and 
$\cos(q', \mu) = \frac{q' \cdot \mu}{\|q'\| \|\mu\|}$ 
is the cosine similarity between the predicted and target embeddings.
We enforce predictions $\vec{\mu}$ to be L2-normalized.

\subsection{Alignment Network}
\label{sec:ap-alignment}

\begin{figure}[t]
    \centering
    \vspace{2mm}
    \includegraphics[width=\linewidth]{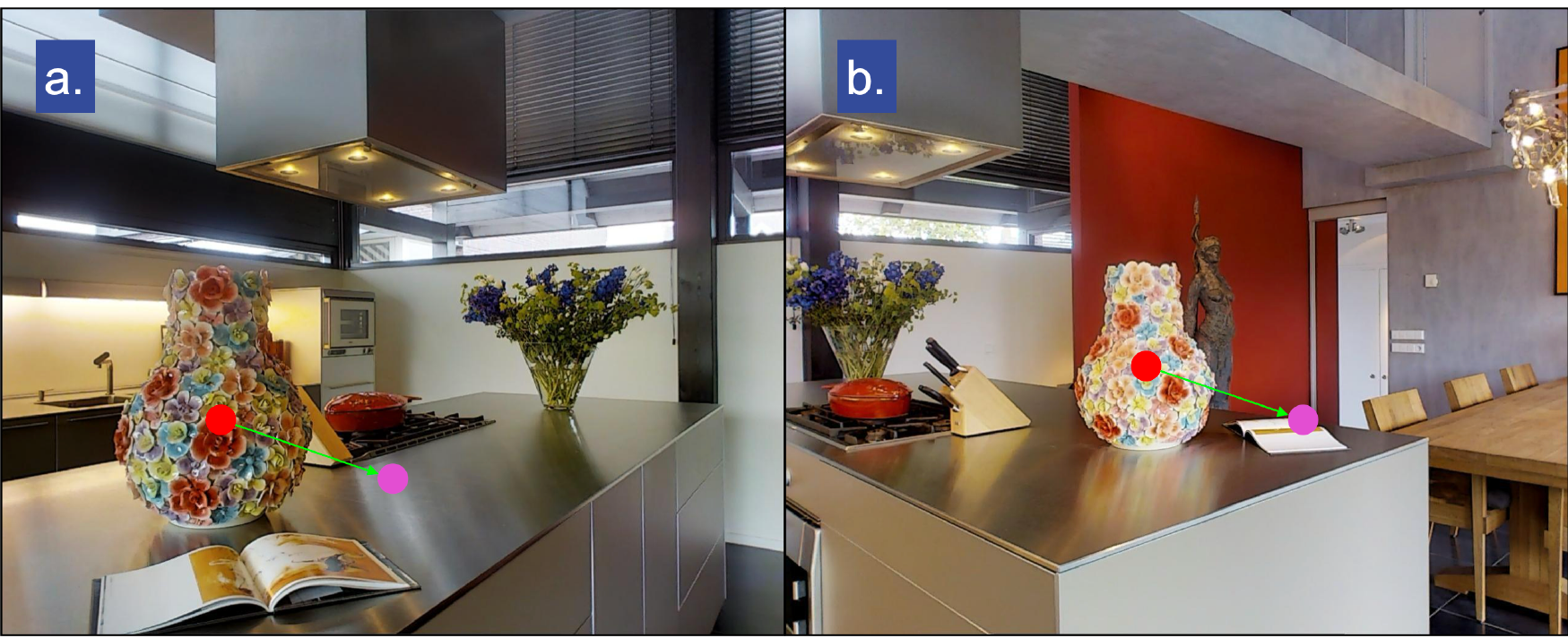}
    \includegraphics[width=\linewidth]{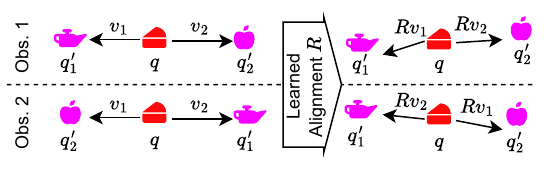}
    \caption{Depiction of the ambiguity problem in the data. Observing the same query feature (red dot), from two different locations can lead to contradictory target features (pink dot) given the same offset vector. In the schematic, we illustrate the effect of the learned alignment on this problem. In two separate observation instances, the alignment $R$ rotates the observations such that the query contradiction is resolved.}
    \label{fig:ambiguity_problem}
\end{figure}

A fundamental challenge of our goal is that the training data can be highly ambiguous. Given that $\vec{v}$ is relative, and our model is not given some constant frame of reference, it is feasible to produce contradictory data points by observing the same scene from two different angles (we illustrate this problem in \cref{fig:ambiguity_problem}). As a remedy, one could curate or filter the dataset to avoid exhibiting the problem. However, this would likely strongly bias the inference of the model towards an artificial data distribution.

We propose an alternative approach: learned dataset decomposition. During training, we introduce an auxiliary network $g$, which is allowed to observe the full training triple. Given such a triple, $g$ outputs a rotation vector $r \in \mathfrak{se3}$ as
\[
    g : \R^E \times \R^3 \times \R^E \rightarrow \mathfrak{se3}.
\]
Using the Rodrigues formula for rotation
\[
\begin{aligned}
    R(\vec{r}, \vec{v}) &= \vec{v} \cos \norm{r} + (\hat{\vec{r}} \times \vec{v}) \sin \norm{\vec{r}} \\ &+ \hat{\vec{r}} (\hat{\vec{r}}^T\vec{v})(1-\cos \norm{\vec{r}}),
\end{aligned}
\]
with $\hat{\vec{r}}$ being the normalized rotation vector $\vec{r}$, we augment the forward pass in our training case as
\[
    \hat{f}(\vec{q}, \vec{v}, \vec{q'}) = f\left(\vec{q}, R\left(g(\vec{q}, \vec{v}, \vec{q'}), \vec{v}\right), \vec{q'}\right).
\]
We can view the final architecture in \cref{fig:model}
This formulation enables the model to learn a suitable data decomposition during training using only rotations. While this removes the ability to infer precise relative locations of objects, it preserves distances, which can be exploited for navigation tasks.
We implement $g$ as an MLP with 2 layers and 256 hidden units.

\subsection{Search Agent}
\label{sec:ap-agent}

We define a navigation agent $A$ tasked with finding a target object $o$ in $T$ discrete navigation steps, where a step consists of moving to a viewpoint in the environment's connectivity graph. At each step, the agent maintains an accumulated semantic point cloud of points
$\{(p_i, q_i)\}_{i=1}^N$ where $p_i \in \mathbb{R}^3$ and $q_i \in \mathbb{R}^E$ 
are the 3D position and feature embedding of each observed point respectively, obtained as explained in \cref{sec:training}. 
We define a target object $o$ for which we obtain a text embedding $q_o \in \mathbb{R}^E$ either directly using CLIP~\cite{radford2021learning}, or by mapping a CLIP embedding to the DINO feature space using Talk2DINO~\cite{barsellotti2025talking}. The agent selects a 
target point $p^* \in \mathbb{R}^3$ in the accumulated point cloud (among unvisited points) and 
navigates to the nearest viewpoint to $p^*$ in the connectivity graph. After navigating to a viewpoint, we mark it as $visited$ together with all points within a radius of 1.5 meters.

To select $p^*$, first, the agent checks whether any observed point is sufficiently similar to the target:
\begin{equation}
    s^* = \max_{i} \, q_i^\top q_o
\end{equation}
We define an exploitation threshold $\tau$, above which $s^* > \tau$ the agent follows the most similar point directly:
\begin{equation}
    p^* = p_i : \argmax_i q_i^\top q_o
\end{equation}
Otherwise, the agent queries \ours{} with $q_o$ over a sphere $\mathcal{S}_r$ 
of uniformly sampled points at radius $r$, obtaining predicted embeddings 
$\{\mu_k\}_{k=1}^{|\mathcal{S}_r|}$ via:
\begin{equation}
\label{field_inference}
    \mu_k, \sigma^2_k = f_\theta(q_o,\, s_k), \quad s_k \in \mathcal{S}_r
\end{equation}
We cluster the predicted embeddings $\mu_k$ into $K$ clusters using $K$-means, 
yielding the \emph{field clustering} $\mathcal{C}_F$. The observed scene 
is then partitioned into spatial cells $\{\mathcal{X}_j\}$, and the embeddings 
within each cell are likewise clustered into $K$ clusters, yielding a 
\emph{scene clustering} $\mathcal{C}_j$ per cell. We mark a cell as $visited$ if at least 80\% of its points are $visited$. The agent selects the 
unvisited cell $\mathcal{X}_{j^*}$ whose scene clustering best matches 
the field clustering under the distance $\mathcal{D}$:
\begin{equation}
    j^* = \argmin_{j \in \text{unvisited}} \, \mathcal{D}(\mathcal{C}_j,\, \mathcal{C}_F).
\end{equation}
The distance $\mathcal{D}$ is the Angular Wasserstein distance of the distributions, which we define in \cref{eq:wasserstein-distance}.
Finally, among all unvisited points belonging to cell $\mathcal{X}_{j^*}$, the agent navigates to the point closest to its current position.

The scoring procedure above operates at a single spatial scale. To handle 
cases where no sufficiently good match is found at the current scale, we 
perform an incremental expansion of the spatial context. We define $L$ radii 
$r_1 < r_2 < \cdots < r_L$. For each level of context $l$, we perform field inference (\cref{field_inference})
using radius $r_l$, obtaining predicted embeddings over the corresponding 
sphere and clustering the predicted means into a field clustering 
$\mathcal{C}_F^l$. This yields a hierarchy of predicted fields encoding 
feature neighborhoods at progressively larger spatial scales.

At each navigation step $t$, the agent evaluates all levels and maintains 
a running best matching cost $c_t^l$ for each level $l$, initialized as 
$c_0^l = \infty$ and updated as:
\begin{equation}
    c_t^l = \min\left(c_{t-1}^l,\ \min_{j \in \text{unvisited}} 
    \mathcal{D}(\mathcal{C}_j,\, \mathcal{C}_F^l)\right).
\end{equation}
At each step, the agent evaluates levels in order, committing to the 
first level $\ell$ at which the current best match represents an 
improvement over the stored cost:
\begin{equation}
    \ell^* = \min \left\{ l \,:\, 
    \min_{j \in \text{unvisited}} \mathcal{D}(\mathcal{C}_j,\, \mathcal{C}_F^l) 
    \leq c_{t-1}^l + \epsilon 
    \right\},
\end{equation}
where $\epsilon$ is a small tolerance. If no level satisfies this condition, the agent falls back to the cell with the lowest cost across all levels:
\begin{equation}
    j^* = \argmin_{j \in \text{unvisited},\, l \in \{1,\dots,L\}} \, 
    \mathcal{D}(\mathcal{C}_j,\, \mathcal{C}_F^l).
\end{equation}

\section{Data Generation \& Training}
\label{sec:training}

To train \ours{} $f$ we use a subset of the Matterport3D~\cite{chang2017matterport3d} 
dataset, a collection of 90 buildings of which we use 20 for training and 6 for validation. For the test of the search agent, we use 20 separate buildings. Each building 
comprises RGB-D panoramic views, along with camera poses.
For each panoramic viewpoint, we encode its 18 RGB images with a pre-trained feature extractor, DINOv2~\cite{oquab2023dinov2}, obtaining L2-normalized patch embeddings of shape 
$H \times W \times E$ per image, with $H=W=37$. We resize the corresponding depth images to the same spatial 
resolution and reconstruct a pointcloud in the local coordinate frame of the viewpoint center, assigning each 3D point its corresponding embedding.
This process is performed once across all $\sim3000$ viewpoints, and the result is stored.
At training time, given a viewpoint, we sample $N = 200$ points 
$\{p_i\}_{i=1}^N \subset \mathbb{R}^3$ with corresponding embeddings 
$\{q_i\}_{i=1}^N \subset \mathbb{R}^d$ and construct training triplets:
\begin{equation}
    \mathcal{D} = \left\{ \left( q_i,\ \vec{v}_{ij},\ q_j \right) 
    \mid i, j \in \{1, \dots, N\} \right\},
\end{equation}
where $\vec{v}_{ij} = p_j - p_i \in \mathbb{R}^3$, yielding $N^2$ 
training triplets per viewpoint. We then minimize $\mathcal{L}$ as defined 
in \cref{eq:loss} over all triplets. To improve generalization, we augment the training data by applying a small 
random spatial perturbation to $\vec{v}$ and a full random rotation, encouraging 
the alignment network to learn a robust canonical frame rather than overfitting 
to the specific geometry of the training scenes.

\section{Experimental Evaluation}
We divide the evaluation of our method into two parts. First, we present the predictive capabilities of \ours{} by examining its ability to predict feature neighborhoods in unseen scenes.
In the second part, we present the results for the object search agent by comparing it to different baselines in the Matterport simulator.

\subsection{Evaluation of Predictive Power}
\label{sec:eval-prediction}
\begin{figure*}[t]
    \centering
    \vspace{2mm}
    \includegraphics[width=0.95\textwidth]{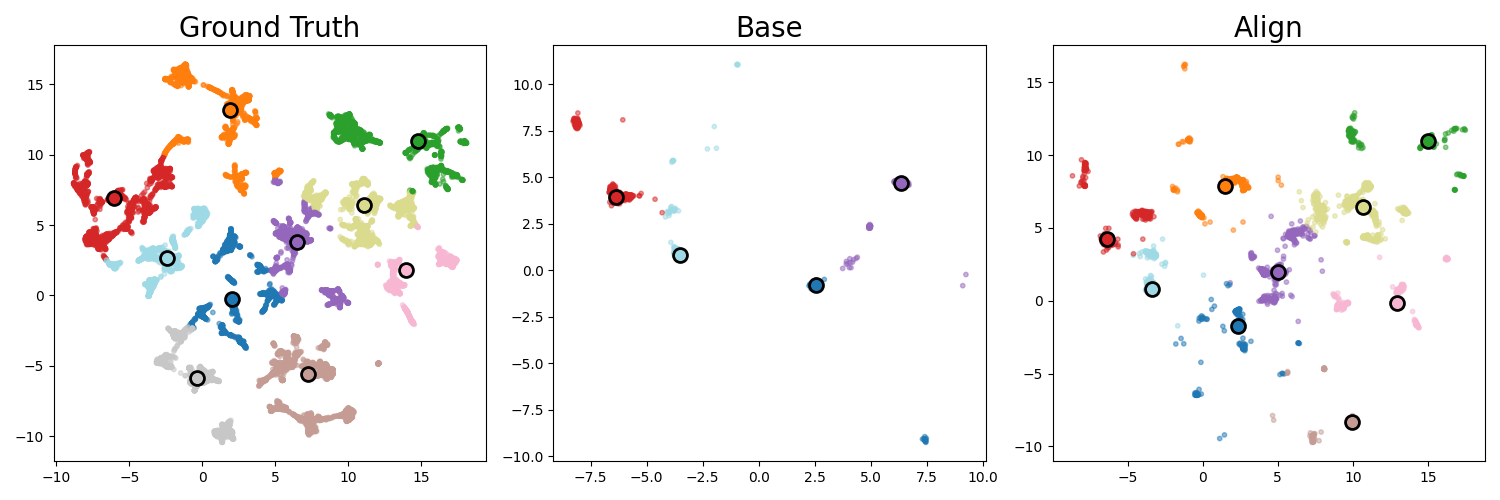}
    \caption{UMAP visualization of ground truth clusters embeddings (left), \textit{base} model predictions (center), and \textit{aligned} model predictions (right). Cluster centroids are marked as circled points. The \textit{base} model exhibits severe mode collapse, whereas the \textit{aligned} model retains substantially more of the semantic diversity around the query embedding. This demonstrates the effectiveness of the Alignment Network at resolving conflicts in the training data.}
    \label{fig:umap}
    \vspace{-3mm}
\end{figure*}

\begin{figure}
    \centering
    \includegraphics[width=\linewidth]{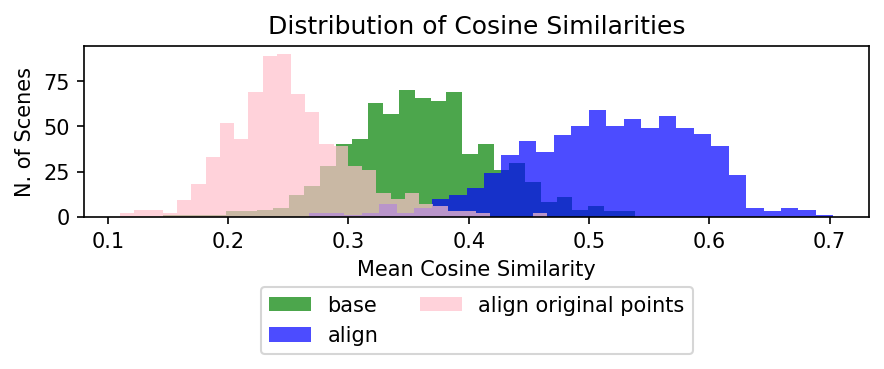}
    \caption{Distribution of pointwise cosine similarities between ground truth and predicted embeddings. \textit{base}: predictor trained without Alignment Network. \textit{aligned}: predictor trained with Alignment Network, tested with positions rotated into canonical frame. \textit{aligned original}: same \textit{aligned} model tested with unrotated scene positions, showing frame-dependency of the learned predictions.}
    \label{fig:cos_sim}
\end{figure}
Before deploying our trained \ours-model, we examine whether it meaningfully captures feature neighborhoods.
We do so by studying its inferences on $6$ unseen buildings comprising $730$ scenes where each scene is a $360^\circ$ viewpoint.

\begin{figure*}[t]
    \centering
    \vspace{2mm}
    \includegraphics[width=\textwidth]{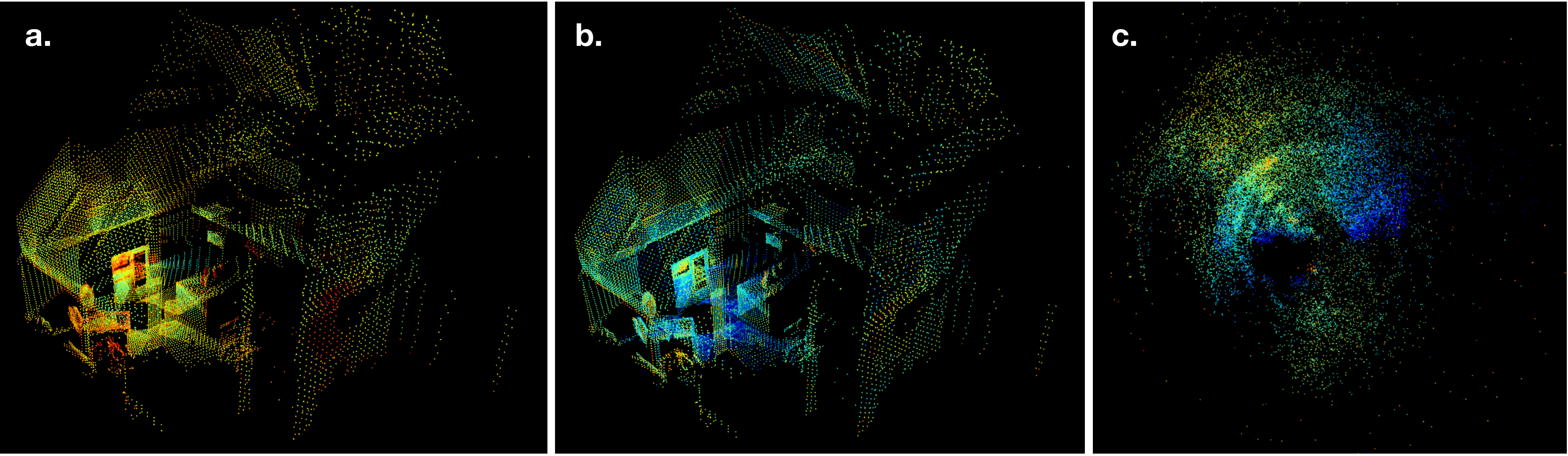}
    \caption{Qualitative impression of similarity difference of predicted features given base model (a) and aligned model (b). Taking a query embedding from a validation scene and querying the model for embeddings of all other points, we observe a stark difference in the semantic accuracy of the predictions. We visualize this here as a heat map (blue = similar, red = dissimilar). In (c), we visualize the impact the rotation predicted by the Alignment Network has on the scene.}
    \label{fig:similarities}
    \vspace{-5mm}
\end{figure*}

\textbf{Pointwise-based evaluation}: 
First, we build the set of evaluation tuples in the same way as done for training in \cref{sec:training}. We then compute the cosine similarity between $q$ (GT) embeddings and  $q'$ (predicted) embeddings for unseen scenes. The results are shown in \cref{fig:cos_sim}. 
The model trained with the Alignment Network achieves higher similarity compared to the basic predictor (\textit{base}), but shows decreased performance when alignment is not employed during inference (\textit{aligned original}), as the model learned to predict embeddings in a canonical frame defined by the alignment network. We present a qualitative result in \cref{fig:similarities}, in which we selected one validation scene, picked one image embedding, and passed it to both predictors, along with the 3D positions relative to the query. We then plotted the pointwise similarity between GT and predictions for the \textit{base} predictor (trained without alignment) (a) and the \textit{aligned} predictor (b), and showed what the \textit{aligned} predictor sees after the Alignment Network rotates the input points (c).
However, such point-wise predictions cannot estimate whether the network captured a meaningful distribution of a feature's semantic neighborhood. 

\textbf{Clustering-based evaluation}: 
For each validation scene, we sample a query position and extract its 
embedding $q \in \mathbb{R}^E$ from the scene. To ensure meaningful and 
diverse queries (avoiding featureless surfaces like walls), we compute the 
mean embedding of the scene and sample randomly from embeddings in the top 
20\% most dissimilar to the mean. We then extract ground truth embeddings 
$\{e_i^{\text{GT}}\}_{i=1}^N$ from all 3D points within a 3-meter radius 
of the query position. We cluster these embeddings into 
$K$ clusters using $K$-means with cosine similarity, obtaining the ground 
truth clustering:
\[
    \mathcal{C}_{\text{GT}} = \{c_1^{\text{GT}}, \ldots, c_K^{\text{GT}}\},
\]
where each $c_k^{\text{GT}}$ is a cluster such that 
$\sum_{k=1}^K |c_k^{\text{GT}}| = N$.

To obtain predicted embeddings, we pass $q$ to \ours{} along with the 
relative positions $\vec{v}$ corresponding to the ground truth points, 
obtaining predictions $\mu_k^P = f_\theta(q, \vec{v})$. We cluster these 
into $K$ clusters to obtain the predicted clustering:
\[
    \mathcal{C}_{P} = \{c_1^{P}, \ldots, c_K^{P}\}.
\]

For each cluster $k$ with centroid $\mu_k$ and member embeddings 
$\{e_i\}_{i \in c_k}$, we compute the angular standard deviation:
\[
    \sigma_k = \sqrt{\frac{1}{|c_k|} \sum_{i \in c_k} 
    \left(\cos^{-1}(\mu_k^\top e_i)\right)^2}
\]

To compare $\mathcal{C}_{\text{GT}}$ and $\mathcal{C}_{P}$, we find the 
optimal cluster correspondence $\pi^*$ using the Hungarian algorithm with 
angular centroid distances as costs. We then compute this metric, inspired by the Wasserstein distance adapted for $\ell_2$-normalized embeddings:
\[\footnotesize
    \label{eq:wasserstein-distance}
    \mathcal{D}(\mathcal{C}_{\text{GT}}, \mathcal{C}_{P}) = \frac{1}{K} 
    \sum_{k=1}^{K} \cos^{-1}\!\left(\mu_k^{\text{GT}} \cdot 
    \mu_{\pi^*(k)}^{P}\right) + 
    \left|\sigma_k^{\text{GT}} - \sigma_{\pi^*(k)}^{P}\right|
\]

This captures both centroid alignment (first term) and spread consistency 
(second term). Lower values indicate better preservation of the semantic 
distribution structure.

\textbf{Baselines}: We compare our models against two baselines: (1) \textit{random}, which generates random unit vectors, and (2) \textit{mean}, which predicts solely the mean embedding of the ground truth scene. All embeddings are clustered.

\begin{figure}
    \centering
    \includegraphics[width=\linewidth]{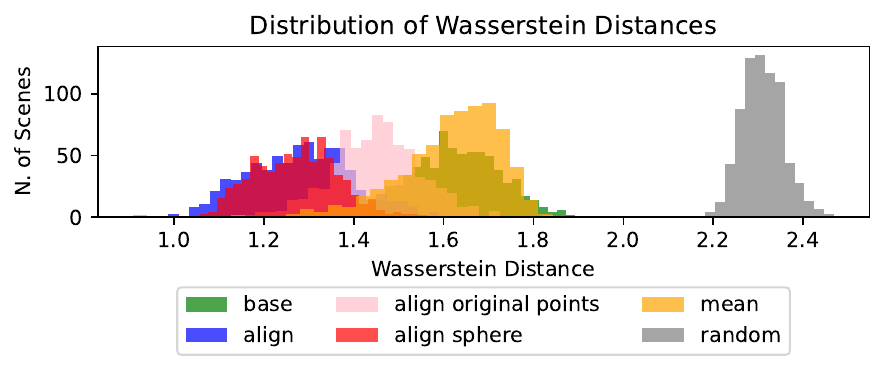}
    \caption{Cluster-based distance between ground truth and predicted embedding distributions. Lower values indicate better preservation of semantic structure. \textit{base}: predictor without alignment. \textit{Aligned}: predictor with Alignment Network using rotated input positions. \textit{Aligned original}: same model using unrotated scene positions. \textit{Aligned sphere}: same model with uniformly sampled spherical positions. \textbf{Baselines}: \textit{random} (random embeddings) and \textit{mean} (all embeddings set to scene mean).}
    \label{fig:wasserstein}
\end{figure}

\textbf{Results}: We evaluate four model configurations with $K = 10$ clusters, shown in \cref{fig:wasserstein}. The predictor trained with the Alignment Network (\textit{aligned}) achieves a substantially lower distance than the basic predictor (\textit{base}) trained without alignment and significantly outperforms both baselines. Notably, the \textit{aligned} model shows consistent performance whether using: (a) original scene points rotated by the Alignment Network (\textit{aligned}), or (b) points uniformly sampled on a 3-meter sphere matching the size of the $\GT$ sample (\textit{aligned sphere}). This demonstrates that the model successfully generalizes beyond the training distribution and can predict semantic neighborhoods from arbitrary spatial queries without requiring the alignment network during inference. Given this result, we query the model for feature neighborhoods $\vec{\mu_e}$ close to a query embedding in a specified radius $r$  by calling $\vec{\mu_e}, \sigma_e = f(q, S)$  where $S$ is a sphere of radius $r$. This makes a general prediction, directly applicable to spatial-semantic search tasks without requiring scene-specific calibration.

\subsection{Object Search}
\label{sec:eval-object-search}

We perform our evaluation on the Matterport3D dataset~\cite{chang2017matterport3d}, which provides a navigation graph for agent movement. Using this graph allows us to focus on the effectiveness of search strategies while minimizing confounding factors from low-level movement or collision dynamics that arise in continuous physics-based environments such as HM3D~\cite{ramakrishnan2021hm3d} or RoboTHOR~\cite{deitke2020robothor}.

\textbf{Experimental Setup}: 
We select 20 buildings ($B$), not belonging to the model's training set, from the Matterport3D dataset, which comprise mainly large multi-floor houses, and defined 24 target objects ($O$) (e.g. cup, bed, kettle...) to search 
for. Each building is composed of a set of scene viewpoints ($S$), connected 
by the Matterport3D simulator's navigation graph. We define a challenge as the 
tuple $c = (b, s, o)$, where $b \in B$ is a 
building, $s \in S$ is the starting scene, and $o \in O$ is the text identifying the target 
object.
For each building $b$ we crafted 5 challenges by selecting 5 different 
starting scenes and 5 target objects such that object diversity is maximized 
across the same building. To ensure that each target object is present in its 
assigned building, we used object labels from the Matterport3D annotations and 
further verified them manually, yielding 100 challenges in total.

All agents operate in the Matterport3D simulator and have access to the full 
connectivity graph of the building as well as perfect 6-DoF poses at every 
step.
At each step, an agent receives an RGB image and a depth image. Using 
the depth image and the known pose, each depth observation is unprojected into 
3D space and merged into a building-level point cloud that accumulates across 
steps. Each 3D point is assigned a VLM embedding (DINOv2 or CLIP) extracted from the 
corresponding RGB patch, producing an incrementally growing semantic point 
cloud. Agents use this shared representation to reason about the environment. 
Navigation actions are discrete. At each step, the agent can select one of the frontier viewpoints in the connectivity graph to navigate to. Agents are 
given a maximum of $T=100$ steps, where a step is a move between viewpoints (rotations are not counted). A challenge is deemed a success if the agent 
reaches a labeled goal viewpoint or if the cosine similarity between the 
agent's current observation embedding and the target query embedding exceeds a 
threshold $\tau_{stop}=0.55$, and a failure otherwise. Each time an agent moves to a new unvisited viewpoint, it does a 360° spin. We subdivide the constructed pointcloud into cells of 3 meters side length.

 \textbf{\ours{}} for our agent (\cref{sec:ap-agent}) we used the following configuration: number of clusters $K = 10$, threshold when to follow the max similar point to the query $\tau = 0.3$, tolerance $\epsilon = 0.05$, max spatial context expansion level $L=2$  with radii $r_1 = 0.1$ and $r_2 = 1.0$. We ablate over the choice of $\tau$ and share the results of the ablation in \cref{tab:tau_ablation}.

\textbf{Baselines}:
We compare against the following baselines: 

\noindent\textit{Query Follower} greedily navigates toward the unvisited viewpoint whose current observation embedding is most similar to the target query, without any predictive model.
We deploy this strategy using both CLIP and DINOv2 features.

\noindent\textit{Random} chooses any unvisited viewpoint to navigate to.

\noindent\textit{BFS} and \textit{DFS} perform blind breadth-first and depth-first traversal of the connectivity graph respectively, using the query embedding only to detect success, serving as uninformed search baselines. 

\noindent\textit{CLIP on Wheels} (CoW)~\cite{gadre2023cows} uses CLIP as an encoder and scores each frontier point by computing the similarity between the scene point embedding and the query embedding. If the score exceeds a threshold, the agent navigates to the corresponding frontier, otherwise it performs a random exploration step.

\noindent\textit{Human Baseline}: 
To contextualize agent performance, we collected human navigation data on the same challenges.
We collected data from 15 participants, each of whom completed 20 challenges.
Participants were shown the same RGB observations available to the agents and instructed to find the target object as efficiently as possible.
The participants never encountered the same environment twice, thereby eliminating learning effects that the agents could not experience. 

\textbf{Evaluation Metrics}: 
To evaluate performance on the challenges, we employ two standard navigation metrics commonly used in the literature~\cite{batra2020objectnav}.

\noindent\textit{Success Rate (SR)}: Fraction of challenges in which the agent found the object.

\noindent\textit{Success weighted by inverse path length (SPL)}: 
\[
\text{SPL} = \frac{1}{\cardinality{C}} \sum_{i=1}^{\cardinality{C}} S_i \cdot \frac{\ell_i}{\max(p_i, \ell_i)},
\]
where $\ell_i$ denotes the optimal path length to address challenge $i$, $p_i$ denotes the path length produced by the agent, and $S_i$ indicates binary success.

\begin{figure*}[t]
    \centering
    \vspace{2mm}
    \includegraphics[width=\textwidth]{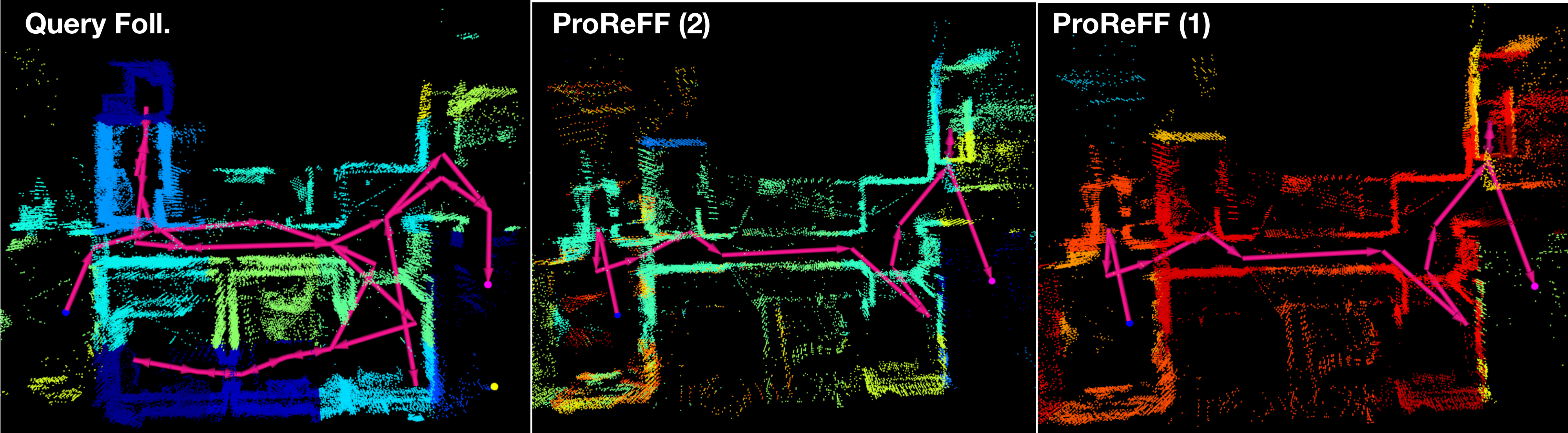}
    \caption{We compare Query Follower against the ProReFF agent and show the navigation path of both (pink arrows). Agents start at the blue circle, in the first picture the Query Follower travels to the goal (pink circle) by following cosine similarities to the query which we draw on the pointcloud as a heatmap where blue = similar, red = dissimilar. The second and third pictures show the ProReFF Wasserstein distance scores for the feature field at the second level of context expansion (2, more coarse) and the first level of expansion (1, near the query, so most of the points are very dissimilar).}
    \label{fig:agent_path}
    \vspace{-2mm}
\end{figure*}

\begin{table}[t]
\centering
\caption{Object search results on MatterSim environments ($N\!=\!100$ episodes).
SR = Success Rate; SPL = Success weighted by Path Length (higher is better for both).
}
\label{tab:search_results}
\hspace{-2mm}
\begin{tabular}{lcccccc}
  \toprule
  & \multicolumn{2}{c}{Overall}
  & \multicolumn{2}{c}{Single-Floor}
  & \multicolumn{2}{c}{Multi-Floor} \\
  \cmidrule(lr){2-3}\cmidrule(lr){4-5}\cmidrule(lr){6-7}
  Method & SR & SPL & SR & SPL & SR & SPL \\
  \midrule
  Random & 0.45 & 0.05 & 0.6 & 0.08 & 0.3 & 0.03 \\
  BFS            & \textbf{0.95} & 0.25 & \textbf{0.98} & 0.27 & 0.92          & 0.23 \\
  DFS            & 0.90          & 0.35 & 0.92          & 0.41 & 0.88          & 0.29 \\
  CoW~\cite{gadre2023cows}    & 0.78          & 0.30 & 0.80          & 0.35 & 0.76          & 0.24 \\
  Query Follower (CLIP) & 0.84 & 0.42 & 0.94 & 0.55 & 0.74 & 0.30 \\
  Query Follower (DINO)& 0.86          & 0.44 & \textbf{0.98} & 0.55 & 0.74          & 0.34 \\
  \ours{} (Ours) & 0.94 & \textbf{0.53} & 0.96 & \textbf{0.63} & \textbf{0.92} & \textbf{0.42} \\
  \midrule
  Mean & 0.82 & 0.33 & 0.88 & 0.41 & 0.75 & 0.26 \\
  \midrule
  \makecell[l]{Human Participants \\ ($N=15$)} & 0.95 & 0.66 & 0.95 & 0.69 & 0.96 & 0.62 \\
  \bottomrule
  \end{tabular}
  \vspace{-5mm}
\end{table}

\textbf{Results}: 
We present the results of the navigation experiments in table~\cref{tab:search_results}.
The overall average success rate lies at $82\%$, with an average SPL of $0.33$. Upon inspection of the results, we note a difference in agent performance when tasked with searching for an object located on another floor. We thus also report the performances for the single-floor and multi-floor challenges separately. 

Given that success rates are generally high, except for the random agent, we focus primarily on the difference in SPL. While BFS achieves the best overall success, \ours{} achieves a similarly high success while also being twice as efficient.
Notably, the \emph{Query Follower} agents also exhibit strong performance. While their success rates are an average of $10$ points lower than \ours{}, they do rank second and third in SPL after \ours{}. 
The CoW agent ranks second lowest on both success rate and SPL. We find this to be the case because it is a mixture of the Random agent and the Query Followers, as it follows embedding similarity only when it exceeds a certain threshold.
All agents show a decrease in SPL in the Multi-Floor scenario compared to the single-floor, but \ours{} remains the most robust.
We find that having to pass a staircase creates a bottleneck in the navigation heuristic for the less context-aware agents. \ours{}, on the other hand, is able to use the semantic context provided by the learned field to consider the stairwell as a worthwhile exploration direction.
The feature fields increase the semantic context, enabling the agent to make informed decisions even in more challenging scenarios, whereas the simple query follower has a smaller, more local semantic context, hence working much better on single-floors.
We observe in \cref{fig:agent_path} that the Wasserstein scores assigned to the point cloud effectively guide the agent toward regions near the goal (green-blue), while discarding less promising areas (yellow-red).
Our human evaluation shows that humans can solve the challenges with equal success and efficiency, regardless of whether they are single- or multi-floor. As expected, our participants achieved the highest success and SPL overall, but surprisingly, the SPL was $0.66$ on average. Given that they can be considered experts at navigating unfamiliar human environments, we infer that the artificial agents achieve $80\%$ of expert performance on average.

We note in our ablation of $\tau$ in \cref{tab:tau_ablation}, that this choice is critical for agent performance. While the overall performance remains competitive over our baselines, further study is needed to determine the specificity of $\tau$ to the environment, which is in this case Human residences.

In our supplementary video, we share a brief qualitative impression of the different agent's behaviors.

\begin{table}[t]
\centering
\caption{Ablation on the threshold parameter $\tau$ for \ours{}.}
\label{tab:tau_ablation}
\begin{tabular}{lcccccc}
\toprule
& \multicolumn{2}{c}{Overall}
& \multicolumn{2}{c}{Single-Floor}
& \multicolumn{2}{c}{Multi-Floor} \\
\cmidrule(lr){2-3}\cmidrule(lr){4-5}\cmidrule(lr){6-7}
$\tau$ & SR & SPL & SR & SPL & SR & SPL \\
\midrule
0.20 & 0.91 & 0.46 & 0.96 & 0.59 & 0.88 & 0.34 \\
0.25 & 0.91 & 0.48 & 0.96 & 0.60 & 0.86 & 0.35 \\
\textbf{0.30} & \textbf{0.94} & \textbf{0.53} & 0.96 & \textbf{0.63} & \textbf{0.92} & 0.42 \\
0.35 & 0.92 & 0.47 & 0.96 & 0.49 & 0.88 & \textbf{0.45} \\
\bottomrule
\end{tabular}
\vspace{-5mm}
\end{table}

\section{Conclusion}
\label{sec:conclusion}

In this work, we presented \ours{} a self-supervised method for training a predictor for object co-occurrences on pre-trained visual features. Different from object-centric methods, \ours{} do not require explicit labels of objects, but can be trained from RGB-D observations. We also proposed an agent exploiting these relative feature fields to enable a downstream object-search task.
In our evaluation, we studied the capabilities of \ours{} to determine the extent to which it learns meaningful co-occurrences of features in natural scenes. Here, we demonstrated especially the utility of the learned data decomposition, which we introduced into the training process. We demonstrate that the model trained with this decomposition can infer a diverse feature neighborhood that matches the ground truth feature distribution.
In evaluating our navigation agent, we tested it against other baselines, some of which also use VLM features, on $100$ navigation challenges in the Matterport 3D environment.
Contrasting the performance of artificial methods with that of humans solving the same task revealed an upper bound on performance to be expected from artificial agents. Measured by this bound, our \ours{} agent achieves $80\%$ of human performance.

Our evaluations have also raised questions that require further inquiry. Two of our baselines, which simply follow feature similarities, perform quite similarly to our more informed agent in single-floor cases.
This raises the question of how much co-occurrence information is already contained in the feature spaces of VLMs, such as DINOv2 and CLIP.
We suspect that the Vision-Transformer's self-attention mechanism and VLMs trained on natural images yield features that already encode information about the typical local neighborhood of an object.
However, as these priors only reflect on the local neighborhood, they cannot capture the full 3D spatial structure of Multi-Floor environments, in which \ours{} proves its utility.
For future research, we aim to study the depth of these local neighborhoods more closely to understand if they can be exploited more efficiently. In addition, we would like to combine \ours{} with a mapping strategy and measure its utility in an embodied agent, in both HM3D~\cite{ramakrishnan2021hm3d} and on a real system.

\balance
\bibliographystyle{IEEEtran}
\bibliography{sources}  %

\end{document}